\documentclass[11pt]{article} 

\RequirePackage[OT1]{fontenc}
\RequirePackage{amsthm,amsmath,amsfonts}
\RequirePackage[colorlinks]{hyperref}
\RequirePackage{hypernat}
\RequirePackage{epsfig, graphicx, color}
\RequirePackage{times}
\RequirePackage{latexsym}
\RequirePackage{psfrag}

\usepackage{fullpage}
\usepackage{epsf,epsfig,subfigure}
\usepackage{fancyheadings}
\usepackage{graphics,graphicx}
\usepackage{enumerate}

\usepackage{epsfig}

\theoremstyle{plain}

\newtheorem{theo}{Theorem}[section]

\newtheorem{lem}{Lemma}[section]
\newtheorem{prop}{Proposition}[section]
\newtheorem{cor}{Corollary}[section]

\theoremstyle{definition} 

\newtheorem{nota}{Notation}[section]
\newtheorem{de}{Definition}[section]
\newtheorem{exa}{Example}[section]
\newtheorem{as}{Assumption}[section]
\newtheorem{alg}{Algorithm}[section]

\newcommand{\btheo}{\begin{theo}}
\newcommand{\bde}{\begin{de}}
\newcommand{\ble}{\begin{lem}}
\newcommand{\bpr}{\begin{prop}}
\newcommand{\bno}{\begin{nota}}
\newcommand{\bex}{\begin{exa}}
\newcommand{\bcor}{\begin{cor}}
\newcommand{\spro}{\begin{proof}}
\newcommand{\bas}{\begin{as}}
\newcommand{\balg}{\begin{alg}}

\newcommand{\etheo}{\end{theo}}
\newcommand{\ede}{\end{de}}
\newcommand{\ele}{\end{lem}}
\newcommand{\epr}{\end{prop}}
\newcommand{\eno}{\end{nota}}
\newcommand{\eex}{\end{exa}}
\newcommand{\ecor}{\end{cor}}
\newcommand{\fpro}{\end{proof}}
\newcommand{\eas}{\end{as}}
\newcommand{\ealg}{\end{alg}}

\theoremstyle{plain}

\newtheorem{theos}{Theorem}
\newtheorem{props}{Proposition}
\newtheorem{lems}{Lemma}
\newtheorem{cors}{Corollary}

\theoremstyle{definition}
\newtheorem{exas}{Example}
\newtheorem{algs}{Algorithm}
\newtheorem{asss}{Asumption}
\newtheorem{defns}{Definition}

\newcommand{\btheos}{\begin{theos}}
\newcommand{\etheos}{\end{theos}}
\newcommand{\bprops}{\begin{props}}
\newcommand{\eprops}{\end{props}}
\newcommand{\bdes}{\begin{defns}}
\newcommand{\edes}{\end{defns}}
\newcommand{\blems}{\begin{lems}}
\newcommand{\elems}{\end{lems}}
\newcommand{\bcors}{\begin{cors}}
\newcommand{\ecors}{\end{cors}}
\newcommand{\bexs}{\begin{exas}}
\newcommand{\eexs}{\end{exas}}
\newcommand{\balgs}{\begin{algs}}
\newcommand{\ealgs}{\end{algs}}
\newcommand{\bass}{\begin{asss}}
\newcommand{\eass}{\end{asss}}

\begin{document}
	\begin{center}
	{\bf{\LARGE{The information geometry of mirror descent}}}

	\vspace*{.1in}
	\begin{tabular}{cc}
	Garvesh Raskutti$^{1}$ & Sayan Mukherjee$^{2,3}$ \\
	\end{tabular}

	\vspace*{.1in}

	\begin{tabular}{c}
	  $^1$ Department of Statistics, University of Wisconsin-Madison\\
	$^2$ Departments of Statistical Science, Computer Science, and
        Mathematics, Duke University\\ 
	$^3$ Institute for Genome Sciences \& Policy, Duke University
	\end{tabular}

	\vspace*{.1in}


	\end{center}

\begin{abstract}
We prove the equivalence of two online learning algorithms, mirror descent and natural gradient descent. Both mirror descent and natural gradient descent are generalizations of online gradient descent when the parameter of interest lies on a non-Euclidean manifold. Natural gradient descent selects the steepest descent along a Riemannian manifold by multiplying the standard gradient by the inverse of the metric tensor. Mirror descent induces non-Euclidean structure by solving iterative optimization problems using different proximity functions. In this paper, we prove that mirror descent induced by a Bregman divergence proximity functions is equivalent to the \emph{natural} gradient descent algorithm on the \emph{dual} Riemannian manifold. We use use techniques from convex analysis and a connections between Riemannian manifolds, Bregman divergences and convexity to prove this result. This equivalence between natural gradient descent and mirror descent,
implies that (1) mirror descent is the steepest descent direction along the Riemannian manifold corresponding to the choice of Bregman divergence; (2) mirror descent with log-likelihood loss applied to parameter estimation in exponential families asymptotically achieves the classical Cram\'er-Rao lower bound, and (3) natural gradient descent for manifolds corresponding to exponential families can be implemented as a first-order method through mirror descent.
\end{abstract}

\section{Introduction}

Recently there has been great interest in online learning both in terms of algorithms as well as in terms of convergence properties.
Given a sequence $\{f_t\}_{t=1}^{\infty}$ of convex differentiable cost functions, $f_t  : \Theta\; \rightarrow  \mathbb{R}$,
with a parameters in a convex set, $ \theta \in \Theta \subset \mathbb{R}^p$, an online learning algorithm predicts a sequence  of
parameters  $\{ \theta_t \}_{t=1}^{\infty}$ which incur a loss $f_t(\theta_t)$ at each iterate $t$. The goal in online learning is to construct a 
sequence that minimizes the \emph{regret} at a time $T$, $\sum_{t=1}^T {f_t(\theta_t)}$.

The most common approach to construct a sequence $\{ \theta_t
\}_{t=1}^{\infty}$ is based on online or stochastic gradient descent. The online gradient descent update is:
\begin{equation}
\label{EqnGradUpdate}
\theta_{t+1} = \theta_t - \alpha_t \nabla f_t(\theta_t),
\end{equation}
where $(\alpha_t)_{t=0}^{\infty}$ denotes a sequence of step-sizes. Gradient descent is the direction of steepest descent if the parameters $\theta_t$ belong to a Euclidean space. However in many applications, parameters lie on non-Euclidean manifolds (e.g. mean parameters for Poisson families, mean parameters for Bernoulli families and other exponential families). In such scenarios gradient descent in the ambient space is not the direction of steepest descent, since the parameter is restricted to a manifold. Consequently generalizations of gradient decent that incorporate non-Euclidean structure have been developed. 

\subsection{Riemannian manifolds and natural gradient descent}

One generalization of gradient descent is \emph{natural gradient descent} developed by Amari~\cite{AmariNatGrad97}. Natural gradient descent assumes the parameter of interest lies on a Riemannian manifold and selects the steepest descent direction along that manifold. Let $(\mathcal{M}, \mathcal{H})$ be a $p$-dimensional Riemannian manifold with metric tensor $\mathcal{H} = (h_{jk})$ and $\mathcal{M} \subset \mathbb{R}^p$. A well-known statistical example of Riemannian manifolds are manifolds induced by the Fisher information of parametric families. In particular given a parametric family $\{p(x; \mu)\}$ where $\mu \in \mathcal{M} \subset \mathbb{R}^p$, let $\{ \mathcal{I}(\mu) \}$ for each $\theta \in \Theta$ denote the $p \times p$ Fisher information matrices. Then $(\mathcal{M}, \mathcal{I}(\mu))$ denotes a $p$-dimensional Riemannian manifold. Table 1 provides examples of statistical manifolds induced by parametric families (see e.g.~\cite{AmariLauritzen87,Cramer46,Rao45} for details).

\begin{table}
\label{TableFamily}  
\begin{center}
\begin{tabular} {| l | l | l |}
\hline
Family & $\mathcal{M}$ & $\mathcal{I}(\mu)$ \\ \hline
$\mathcal{N}(\theta, I_{p \times p})$ & $\mathbb{R}^p$ & $I_{p \times p}$ \\ \hline
$\mbox{Bernoulli}(p)$ & $[0,1]$ & $\frac{1}{p(1-p)}$ \\ \hline
$\mbox{Poisson}(\lambda)$ & $[0, \infty)$ & $\frac{1}{\lambda}$\\
\hline
\end{tabular}
\caption{Statistical manifold examples}
\end{center}
\end{table}
When $\mathcal{I}(\theta) = I_{p \times p}$, the Riemannian manifold corresponds to standard Euclidean space. For a thorough introduction to Riemannian manifolds, see ~\cite{doCarmo}. 

Given a sequence of functions $\{\tilde{f}_t\}_{t=0}^{\infty}$ on the Riemannian manifold $\tilde{f}_t : \mathcal{M} \rightarrow \mathbb{R}$, the \emph{natural} gradient descent step is:
\begin{equation}
\label{EqnNaturalGradient}
\mu_{t+1} = \mu_t - \alpha_t \mathcal{H}^{-1}(\mu_t) \nabla \tilde{f}_t (\mu_t),
\end{equation}
where $\mathcal{H}^{-1}$ is the inverse of the Riemannian metric
$\mathcal{H} = (h_{jk})$ and $\mu$ is the parameter of interest. If $(\mathcal{M}, \mathcal{H}) = (\mathbb{R}^p, I_{p \times p})$, the natural gradient step corresponds to the standard gradient descent step ~\eqref{EqnGradUpdate}. Theorem 1 in ~\cite{AmariNatGrad97} proves
that the natural gradient algorithm steps in the direction of steepest
descent along the Riemannian manifold $(\mathcal{M},
\mathcal{H})$. Hence the name natural gradient descent.

\subsection{Mirror descent with Bregman divergences}

Another generalization of online gradient descent is mirror descent developed by Nemirovski and Yudin~\cite{NemirovskiYudin83}. Mirror descent induces non-Euclidean geometry by re-writing the gradient descent update as an iterative $\ell_2$-penalized optimization problem and selecting a proximity function different from squared $\ell_2$ error. Note that the online gradient descent step~\eqref{EqnGradUpdate} can alternatively be expressed as:
\begin{equation*}
\theta_{t+1} = \arg \min_{\theta \in \Theta} \left\{\langle \theta,
  \nabla f_t (\theta_t) \rangle + \frac{1}{2 \alpha_t} \| \theta - \theta_t \|_2^2 \right\},
\end{equation*}
where $\Theta \subset \mathbb{R}^p$. By re-expressing the stochastic gradient step in this way, ~\cite{NemirovskiYudin83} introduced a generalization of gradient descent as follows: Denote the \emph{proximity} function $\Psi : \mathbb{R}^p \times \mathbb{R}^p \rightarrow \mathbb{R}^+$, strictly convex in the first argument, then define the \emph{mirror} descent step as:
\begin{equation}
\label{EqnMD}
\theta_{t+1} = \arg \min_{\theta \in \Theta} \left\{\langle \theta,
  \nabla f_t (\theta_t) \rangle + \frac{1}{\alpha_t} \Psi(\theta , \theta_t) \right\}.
\end{equation}
Setting $\Psi(\theta, \theta') = \frac{1}{2}\|\theta - \theta'\|_2^2$ yields the standard gradient descent update, hence ~\eqref{EqnMD} is a generalization of online gradient descent.

A standard choice for the proximity function $\Psi$ is the so-called \emph{Bregman divergence} since it corresponds to the Kullback-Leibler divergence for different exponential families. In particular, let $G : \Theta \rightarrow \mathbb{R}$ denote a strictly convex twice-differentiable function, the divergence introduced by ~\cite{Bregman67a} $B_G : \Theta \times \Theta \rightarrow \mathbb{R}^+$ is:
\begin{equation*}
B_G(\theta, \theta') = G(\theta) - G(\theta') - \langle \nabla G (\theta'), \theta - \theta'  \rangle.
\end{equation*}
Bregman divergences are widely used in statistical inference, optimization, machine learning, and information geometry (see e.g.~\cite{AmariCichocki10, Banerjee05}). Letting $\Psi(\cdot, \cdot) = B_G(\cdot,\cdot)$, the mirror descent step defined is:
\begin{equation}
\label{EqnMirrorDescent}
\theta_{t+1} = \arg \min_{\theta} \left\{ \langle \theta, \nabla f_t (\theta_t) \rangle + \frac{1}{\alpha_t} B_G(\theta,\theta_t) \right\}.
\end{equation}
There is a one-to-one correspondence between Bregman divergences and exponential families ~\cite{Banerjee05} which we exploit later when we discuss estimation in exponential families. Examples of $G$, exponential families and the induced Bregman divergences are listed in  Table 2. For a more extensive list, see e.g.~\cite{Banerjee05}. 

\begin{table}
\label{TableBregman}  
\begin{center}
\begin{tabular} {| l | l | l |}
\hline
Family & $G(\theta)$ & $B_G(\theta, \theta')$ \\ \hline
$\mathcal{N}(\theta, I_{p \times p})$ & $\frac{1}{2}\| \theta\|_2^2$ & $\frac{1}{2}\| \theta - \theta'\|_2^2$ \\ \hline
$\mbox{Poisson}(e^{\theta})$ & $\exp(\theta)$ & $\exp(\theta) - \exp(\theta') - \langle \exp(\theta'), \theta - \theta' \rangle$ \\ \hline
$\mbox{Bernoulli}(\frac{1}{1+e^{-\theta}})$ & $\log(1 + \exp(\theta))$ & $\log \biggr ( \frac{1 + e^\theta}{1 + e^{\theta'}}  \biggr) - \langle \frac{e^{\theta'}}{1 + e^{\theta'}}, \theta - \theta' \rangle$\\ 
\hline
\end{tabular}
\caption{Bregman divergence examples}
\end{center}
\end{table}

\subsection{Our contribution}

In this paper, we prove that the mirror descent update with Bregman divergence step ~\eqref{EqnMirrorDescent} is equivalent to the natural gradient step~\eqref{EqnNaturalGradient} along the \emph{dual} Riemannian manifold which we introduce later. The proof of equivalence uses concepts in convex analysis combined with connections between Bregman divergences and Riemannian manifolds developed in~\cite{AmariCichocki10}. Using the equivalence of the two algorithms, we can exploit the desireable properties of both algorithms. In particular natural gradient descent is known to be the direction of steepest descent along a Riemannian manifold and is Fisher efficient for parameter estimation in exponential families, neither of which are known for mirror descent. From an algorithmic perspective, mirror descent is a first-order method whereas natural gradient descent is a second-order method so implementing natural gradient descent using mirror descent has potential algorithmic advantages.

\section{Equivalence through dual co-ordinates}

In this section we prove the equivalence of natural gradient descent~\eqref{EqnNaturalGradient} and mirror descent ~\eqref{EqnMirrorDescent}. The key to the proof involves concepts in convex analysis, in particular the convex conjugate function and connections between Bregman divergences, convex functions and Riemannian manifolds. 

\subsection{Bregman divergences and convex duality}

The concept of convex conjugate functions is central to the main result in the paper. The convex conjugate function for an function $G$ is defined to be:
\begin{equation*}
H(\mu) : = \sup_{\theta \in \Theta} \left\{ \langle \theta, \mu \rangle - G(\theta)  \right\}.
\end{equation*}
If $G$ is lower semi-continuous,  $G$ is the convex conjugate of $H$, implying a dual relationship between $G$ and $H$. 
Further, if we assume $G$ is strictly convex and twice differentiable, then so is $H$. Note also that if $g = \nabla G$ and $h = \nabla H$, $g = h^{-1}$. For additional properties and motivation for the convex conjugate function, see~\cite{Rockafeller}. 

Let $\mu = g(\theta) \in \Phi$ be the point at which the
supremum for the dual function is attained represent the \emph{dual}
co-ordinate system to $\theta$. The dual Bregman divergence $B_H :
\Phi \times \Phi \rightarrow \mathbb{R}^+$ induced by the strictly convex differentiable function $H$ is:
\begin{equation*}
B_H(\mu, \mu') = H(\mu) - H(\mu') - \langle \nabla H(\mu'), \mu- \mu' \rangle.
\end{equation*}  
Using the dual co-ordinate relationship, it is straightforward to show that $B_H(\mu, \mu') = B_G(h(\mu'), h(\mu))$ and $B_G(\theta, \theta') = B_H(g(\theta'), g(\theta))$. Dual functions and Bregman divergences for examples in Table 2 are presented in Table 3. For more examples see ~\cite{Banerjee05}.

\begin{table}
\label{TableBregmanDual}
\begin{center}
\begin{tabular} {| l | l | l |}
\hline
$G(\theta)$ & $H(\mu)$ & $B_H(\mu, \mu')$ \\ \hline
$\frac{1}{2}\| \theta\|_2^2$ & $\frac{1}{2} \| \mu \|_2^2$ &  $\frac{1}{2}\| \mu - \mu'\|_2^2$ \\ \hline
$\exp(\theta)$ & $\langle \mu, \log \mu \rangle - \mu $ & $\mu \log \frac{\mu}{\mu'}$ \\ \hline
$\log(1 + \exp(\theta))$ & $\eta \log \mu + (1-\mu) \log(1 - \mu)$ & $(1 - \mu) \log \biggr ( \frac{1-\mu}{1-\mu'} \biggr ) + \mu \log \frac{\mu}{\mu'}$ \\ 
\hline
\end{tabular}
 \caption{Dual Bregman divergence examples}
\end{center}
\end{table}

\subsection{Bregman divergences and Riemannian manifolds}

Now we explain how every Bregman divergence and its dual induces a
pair of Riemannian manifolds as described in~\cite{AmariCichocki10}. For the Bregman divergence $B_G :
\Theta \times \Theta \rightarrow \mathbb{R}^+$ induced by the convex
function $G$, define the Riemannian metric on $\Theta$, $\mathcal{G} =
\nabla^2 G$ (i.e. the Hessian matrix). Since $G$ is a strictly convex twice differentiable
function, $\nabla^2 G(\theta)$ is a positive definite matrix for all $\theta \in \Theta$.
Hence $B_G(\cdot,\cdot)$ induces the Riemannian manifold $(\Theta, \nabla^2 G)$. Now let $\Phi$ be the image of $\Theta$ under the continuous map $g = \nabla G$. $B_H
: \Phi \times \Phi \rightarrow \mathbb{R}^+$ induces a Riemannian
manifold $(\Phi, \mathcal{H})$, where $\mathcal{H} = \nabla^2 H$.  Let
$(\Theta, \nabla^2 G)$ denote the \emph{primal} Riemannian manifold
and $(\Phi, \nabla^2 H)$ denote the \emph{dual} Riemannian manifold. 

For example, for the Gaussian statistical family defined on Table 1, $\Theta = \Phi = \mathbb{R}^p$ and $\nabla^2 G = \nabla^2 H = I_{p \times p}$ (i.e. the primal and dual manifolds are the same). On the other hand, for the $\mbox{Bernoulli}(p)$ family in Table 1, the mean parameter is $p$ whereas the natural parameter is $\theta = \log p - \log(1-p)$ and $G(\theta) = \log(1 + e^\theta)$. Consquently $(\Theta, \nabla^2 G) = (\mathbb{R}, \frac{e^{-\theta}}{(1+e^{-\theta})^2})$ and $(\Phi, \nabla^2 H) = ([0,1], \frac{1}{p(1-p)})$ which is consistent with Table 1.

\subsection{Main Result}

In this section we present our main result, the equivalence of mirror descent and natural gradient descent. We also discuss consequences and implications.

\btheos
\label{ThmMain}
The mirror descent step~\eqref{EqnMirrorDescent} with Bregman divergence defined by $G$ applied to the sequence of functions $(f_t)_{t=0}^{\infty}$ in the space $\Theta$ is equivalent to the natural gradient step ~\eqref{EqnNaturalGradient} along the dual Riemannian manifold $(\Phi, \nabla^2 H)$. 
\etheos
The proof follows by stating mirror descent in the dual Riemannian manifold and simple applications of the chain rule.

\spro
Recall that the mirror descent update is:
\begin{equation*}
\theta_{t+1} = \arg \min_{\theta} \left\{ \langle \theta, \nabla f_t (\theta_t) \rangle + \frac{1}{\alpha_t} B_G(\theta,\theta_t) \right\}.
\end{equation*}
Finding the minimum by differentiation yields the step: 
\begin{equation*}
g(\theta_{t+1}) = g(\theta_t) - \alpha_t \nabla_{\theta} f_t(\theta_t),
\end{equation*}
where $g = \nabla G$.  In terms of the dual variable $\mu = g(\theta)$ and noting that $\theta = h(\mu) = \nabla H(\mu)$,
\begin{equation*}
\mu_{t+1} = \mu_t - \alpha_t \nabla_{\theta} f_t(h(\mu_t)).
\end{equation*}
Applying the chain rule to $\nabla_{\mu} f_t(h(\mu)) = \nabla_{\mu}
h(\mu)  \nabla_{\theta} f_t(h(\mu))$ implies that 
$$\nabla_{\theta} f_t(h(\mu_t)) = [\nabla_{\mu} h(\mu_t)]^{-1}
\nabla_{\mu} f_t(h(\mu_t)).$$ 
Therefore
\begin{equation*}
\mu_{t+1} = \mu_t - \alpha_t [\nabla^2 H(\mu_t)]^{-1} \nabla_{\mu} f_t(h(\mu_t)),
\end{equation*}
which corresponds to the natural gadient descent step. This completes the proof.
\fpro

\section{Consequences}

In this section, we discuss how this connection directly yields optimal efficiency results for mirror descent and discuss connections to other online algorithm on Riemannian manifolds.

Firstly by Theorem 1 in Amari~\cite{AmariNatGrad97}, natural gradient descent along the Riemannian manifold $(\Phi, \nabla^2 H)$ follows the direction of steepest descent along that manifold. As an immediate consequence, mirror descent with Bregman divergence induced by $G$ follows the direction of steepest descent along the Riemannain manifold $(\Phi, \nabla^2 H)$ where $H$ is the convex conjugate for $G$. As far as we are aware, an interpretation in terms of Riemannian manifolds had not been provided for mirror descent.

Secondly from an algorithmic perspective notice that natural gradient descent is a second-order method since it requires computation of the metric tensor $\nabla^2 H$ whereas mirror descent is a first-order method since it simply requires the derivative of $f$ and $G$ at each step. For many large-scale statistical inference problems first-order methods are preferred since computation of the derivative is significantly less intensive compared to computation of the hessian. Hence using the equivalence of natural gradient and mirror descent, the natural gradient descent can be implemented as a first-order method which has potential computational benefits.  

Next we explain how using existing theoretical results in Amari~\cite{AmariNatGrad97}, we can prove that mirror descent achieves Fisher efficiency.

\subsection{Efficient parameter estimation in exponential families}
\label{SecExpFam}

In this section we exploit the connection between mirror descent and
natural gradient descent to study the efficiency of mirror descent
from a statistical perspective. Prior work on the statistical theory
of mirror descent has largely focussed on regret analysis and we are
not aware of analysis of second-order properties such as statistical efficiency. We will see that Fisher efficiency \cite{Efron75,Fisher25,Rao61} which is an optimality criterion on the covariance of a parameter estimate is an immediate consequence of the equivalence between mirror descent and natural gradient descent.

The statistical problem we consider is parameter estimation in exponential families. Consider a \emph{natural parameter} exponential family with density:
\begin{equation*}
p(y \mid \theta)  = h(y) \exp(\langle \theta, y \rangle - G(\theta) ),
\end{equation*}
where $\theta \in \mathbb{R}^{p}$ and $G: \mathbb{R}^p \rightarrow \mathbb{R}$ is a strictly convex differentiable function. The probability density function can be re-expressed in terms of the Bregman divergence $B_G(\cdot,\cdot)$ as follows:
\begin{equation*}
p(y \mid \theta)  = \tilde{h}(y) \exp( -B_G(\theta, h(y)) ),
\end{equation*} 
where recall that $h = \nabla H$ and $H$ is the conjugate dual function of $G$. The distribution can be expressed in terms of the \emph{mean parameter} $\mu = g(\theta)$ and the dual Bregman divergence $B_H(\cdot,\cdot)$:
\begin{equation*}
p(y \mid \eta)  = \tilde{h}(y) \exp( -B_H(y, \mu) ).
\end{equation*}
As mentioned earlier, there is a one-to-one correspondence between exponential families and Bregman divergence ~\cite{AzouryWarmuth01,Banerjee05}. 

Consider the mirror descent update for the natural parameter $\theta$ with proximty function $B_G(\cdot,\cdot)$ when the function to be minimized is the standard log loss:
\begin{equation*}
f_t(\theta ; y_t) = - \log p(y_t \mid \theta) = B_G(\theta, h(y_t)).
\end{equation*}
Then the mirror descent step is:
\begin{equation}
\label{EqnMDStat}
\theta_{t+1} = \arg \min_{\theta} \biggr \{ \langle \theta, \nabla_{\theta} B_G(\theta, h(y_t)) |_{\theta = \theta_t} \rangle + \frac{1}{\alpha_t}  B_G(\theta, \theta_t) \rangle   \biggr \}.
\end{equation}
Now if we consider the natural gradient descent step for the mean parameter $\mu$, the function to be minimized is again the standard log-loss in the $\mu$ co-ordinates:
\begin{equation*}
\tilde{f}_t(\mu ; y_t) = - \log p(y_t \mid \mu) = B_H(y_t, \mu).
\end{equation*}
Using Theorem~\ref{ThmMain} (or by showing it directly), the natrual gradient step is:
\begin{equation}
\label{EqnNatStat}
\mu_{t+1} = \mu_t - \alpha_t [\nabla^2 H]^{-1} \nabla B_H(y_t, \mu_t).
\end{equation}
A parallel argument holds if the mirror descent step was expressed in terms of the mean parameter and the natural gradient step in terms of the natural parameter. 

Now we use Theorem 2 in ~\cite{AmariNatGrad97} to prove that mirror descent yields an asymptotically Fisher efficient for $\mu$. The Cram\'{e}r-Rao theorem states that any unbiased estimator based on $T$ independent samples $y_1, y_2,...,y_T$ of $\mu$, which we denote by $\widehat{\mu}_T$ satisfies the following lower bound:
\begin{equation*}
\mathbb{E}[(\widehat{\mu}_T - \mu)(\widehat{\mu}_T - \mu)^T] \succeq \frac{1}{T} \nabla^2 H,
\end{equation*}
where $\succeq$ refers to the standard matrix inequality.
A sequence of estimators $(\widehat{\mu}_t)_{t=1}^{\infty}$ is asymptotically Fisher efficient if:
\begin{equation*}
\lim_{T\rightarrow \infty} T\mathbb{E}[(\widehat{\mu}_T - \mu)(\widehat{\mu}_T - \mu)^T] \rightarrow \nabla^2 H.
\end{equation*}
Now by using Theorem 2 in ~\cite{AmariNatGrad97} for natural
gradient descent and the equivalence of natural gradient descent and mirror descent (Theorem~\ref{ThmMain}), it follows that mirror descent is Fisher efficient.

\bcors The mirror descent step applied to the log loss~\eqref{EqnMDStat} with step-sizes $\alpha_t = \frac{1}{t}$ asymptotically achieves the Cram\'{e}r-Rao lower bound.
\ecors
For a more detailed discussion on the statistical properties of natural gradient see~\cite{AmariNatGrad97}. Here we have illustrated how the equivalence between mirror descent with Bregman divergences and natural gradient descent gives second-order optimality properties of mirror descent.

\subsection{Connection to other online methods on Riemannian manifolds}
\label{SecRiemGrad}

The point in using the natural gradient is to the parameter of
interest in the direction of the gradient on the manifold rather than
the gradient in the ambient space. Note however that any
non-infinitesimal step in the direction of the gradient of the
manifold will move one off the manifold, for any curved manifold. This
observation has motivated algorithms ~\cite{Bonnabel11} in which the
update step is constrained to remain on the manifold. 

In this section, we discuss the relation between natural gradient 
descent, mirror descent, and gradient based methods that along a
Riemannian manifold ~\cite{Bonnabel11}. To define the online steepest 
descent step used in ~\cite{Bonnabel11}, we need to define the 
exponential map and differentiation in curved spaces.

The \emph{exponential map} at a point $\mu \in \cal{M}$ 
is a map $\exp_\mu : T_\mu \cal{M} \rightarrow \cal{M}$ where $T_\mu
\cal{M}$ is the tangent space at each point $\mu \in \cal{M}$ (see
e.g.~\cite{doCarmo}). The idea of an exponential map is starting at a 
point $\mu$ with tangent vector $v \in T_\mu$ if one starts at point
$\mu$ and ``flows'' along the manifold in direction $v$ for a fixed (unit)
time interval at coinstant velocity one reaches a new point on the
manifold $\exp_\mu(v)$. This idea is usual stated in terms of geodesic
curves on the manifold, consider the geodesic curve $\gamma : [0,1]
\rightarrow \cal{M}$, with $\gamma(0) = \mu$ and $\dot{\gamma}(0) =
v$, where $v \in T_\mu \cal{M}$ then $\exp_\mu(v) = \gamma(1)$. Again,
in words,  $\exp_\mu(\cdot)$ is the end-point of a curve that lies
along the manifold $\cal{M}$ that begins at $\mu$ with initial
velocity $v = \dot{\gamma}(0)$ that travels one time unit.

Now we define differentiation along a manifold. Let $f : \cal{M} \rightarrow \mathbb{R}$ be a differentiable function on $\cal{M}$. The gradient vector field $\bigtriangledown_{\cal{M}} f $ takes the form $\bigtriangledown_{\cal{M}} f (\mu) = \bigtriangledown_v(f(\exp_\mu(v)) ) |_{v=0}$ noting that $f(\exp_\mu(v))$ is a smooth function on $T_\mu \cal{M}$.

For the sequence of functions $\{f_t\}_{t=0}^{\infty}$ where $f_t : \mathcal{M} \rightarrow \mathbb{R}$ the online steepest descent step analyzed in~\cite{Bonnabel11} is:
\begin{equation}
\label{RiemGDStep}
\mu_{t+1} = \exp_{\mu_t}(- \alpha_t \nabla_{\cal{M}} f_t(\mu_t)).
\end{equation}
The key reason why the update ~\eqref{RiemGDStep} is the standard gradient descent step instead of the natural gradient descent step introduced by Amari is that $\mu_{t+1}$ is always guaranteed to lie on the manifold $\mathcal{M}$ for ~\eqref{RiemGDStep}, but not for the natural gradient descent step. Unfortunately, the exponential map is extremely difficult to evaluate in general since it is the solution of a system of second-order differential equations~\cite{doCarmo}. 

Consequently a standard strategy is to use a computable \emph{retraction} $R_\mu :  T_\mu {\cal{M}} \rightarrow {\mathbb{R}}^p$ of the exponential map which yields the approximate gradient descent step:
\begin{equation}
\label{restRiemGDStep}
\mu_{t+1} = R_{\mu_t}(- \alpha_t \nabla_{\cal{M}} f_t(\mu_t)).
\end{equation}
The retraction $R_{\mu}(v) = \mu +  v$ corresponds to the
first-order Taylor approximation of the exponential map  and yields
the natural gradient descent step in
~\cite{AmariNatGrad97}. Therefore as pointed out in~\cite{Bonnabel11}, natural gradient descent can be cast as an approximation to gradient descent for Riemannian manifolds. Consequently mirror descent can be viewed as an easily computable first-order approximation to steepest descent for any Riemannian manifold induced by a Bregman divergence. 

\section{Discussion}
\label{SecDiscussion}

In this paper we prove that mirror descent with proximity function
$\Psi$ equal to a Bregman divergence is equivalent to the natural
gradint descent algorithm along the dual Riemannian manifold. Based on this equivalence, we use results developed by~\cite{AmariNatGrad97} to conclude that mirror descent is the direction of steepest in the corresponding Riemannian space and for parameter estimation in exponential families with the associated Bregman divergence, mirror descent achieves the Cram\'{e}r-Rao lower bound. Furthermore, this connection proves that the natural gradient step can be implemented as a first-order method using mirror descent which has computational gains for larger datasets.

Following on from this connection, there are a number of interesting and open directions. Firstly, one of the important issues for any online learning algorithm is choice of step-size. Using the connection between mirror descent and natural gradient, it would be interesting to determine whether adaptive choices of step-sizes proposed in~\cite{AmariNatGrad97} that exploit the Riemannian structure can improve performance of mirror descent. It would also be useful to determine a precise characterization of the geometry of mirror descent for other proximity functions such as $\ell_p$-norms and explore links online algorithms such as projected gradient descent.

\subsection*{Acknowledgements}
GR was partially supported by the NSF under Grant DMS-1127914 to the Statistical and Applied Mathematical Sciences Institute. SM was supported by  grants: NIH (Systems Biology): 5P50-GM081883, AFOSR: FA9550-10-1-0436, and NSF CCF-1049290.

\bibliographystyle{plain} 

\bibliography{Oct14Bib.bib}

\end{document}